\documentclass{article}

\usepackage{microtype}
\usepackage{graphicx}
\usepackage{subfigure}
\usepackage{mathtools}
\usepackage{booktabs}
\newcommand{\phrase}[1]{{\tt\small{"#1"}}}
\newtheorem{exmp}{Example}
\usepackage{xcolor}
\usepackage{hyperref}

\usepackage[accepted]{icml2021}

\icmltitlerunning{Powering Effective Climate Communication with a Climate Knowledge Base}

\begin{document}

\twocolumn[
\icmltitle{Powering Effective Climate Communication with a Climate Knowledge Base}

\icmlsetsymbol{equal}{*}

\begin{icmlauthorlist}
\icmlauthor{Kameron B. Rodrigues}{su}
\icmlauthor{Shweta Khushu}{ss}
\icmlauthor{Mukut Mukherjee}{ui}
\icmlauthor{Andrew Banister}{cm}
\icmlauthor{Anthony Hevia}{cm}
\icmlauthor{Sampath Duddu}{hi}
\icmlauthor{Nikita Bhutani}{ml}
\end{icmlauthorlist}

\icmlaffiliation{su}{Stanford University, USA}
\icmlaffiliation{cm}{Climate Mind, USA}
\icmlaffiliation{ml}{Megagon Labs, USA}
\icmlaffiliation{ss}{SkySpecs Inc, USA}
\icmlaffiliation{ui}{University of Illinois, Urbana-Champaign, USA}
\icmlaffiliation{hi}{Hippo Insurance, USA}

\icmlcorrespondingauthor{Kameron B. Rodrigues}{kameron.rodrigues@stanford.edu}

\icmlkeywords{Natural Language Processing, Knowledge Graph, Knowledge Base, Ontology, Climate Change, BERT, Climate Conversation, personal values, Climate Change Awareness}

\vskip 0.3in
]

\printAffiliationsAndNotice{} 

\begin{abstract}
While many accept climate change and its growing impacts, few converse about it well, limiting the adoption speed of societal changes necessary to address it. In order to make effective climate communication easier, we aim to build a system that presents to any individual the climate information predicted to best motivate and inspire them to take action given their unique set of personal values. To alleviate the cold-start problem, the system relies on a knowledge base (ClimateKB) of causes and effects of climate change, and their associations to personal values. Since no such comprehensive ClimateKB exists, we revisit knowledge base construction techniques and build a ClimateKB from free text. We plan to open source the ClimateKB and associated code to encourage future research and applications. 

\end{abstract}

\section{Introduction}
\label{introduction}

Today, climate change is widely recognized as one of the biggest and most threatening global challenges currently facing humanity~\cite{ipccsr15}.

 While 72\% of adult Americans think global warming is happening, there is a lack of acceptance that it personally affects us and a lack of motivation to address it, according to 2020 national public opinion polling~\cite{howe2015geographic, 6americasReview}.

This is demonstrated by how 47\% think global warming will harm them little or not at all, and that 2 out of every 3 Americans rarely or never discuss climate change. Lack of public engagement on climate change for any country can threaten its ability to reach its Nationally Determined Contribution to reduce emissions for the Paris agreement.

Another critical reason for the lack of motivation for action and disconnect with personal risk acceptance are cognitive biases, such as confirmation bias, motivated reasoning, and cultural cognition. Standard communication about climate change must not ignore these biases, and will likely be ineffective if simply assuming a passive, blank-slate audience~\cite{scheufele14,druckman2019evidence,akin2017recap}. 
Several studies suggest that effective and motivating climate communication requires transmission to be shaped based on underlying values, mental and cultural models of the audience and include practical, viable, accessible, and attractive solutions to address it~\cite{carpenter2019cognitive,iyengar2019scientific}. 

\begin{figure}[ht]
    \centering
    \includegraphics[scale=0.75]{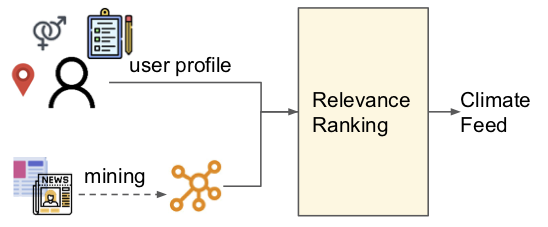}
    \caption{Effective climate communications using ClimateKB and relevance characteristics from a user's profile.}
    \label{fig:pipeliine}
\end{figure}

Our goal is to make climate communication more effective with a recommendation system informed by personal values and powered by climate change concepts mined from current, reputable, and truthful climate news articles. Inspired by framing theory~\cite{chong2007framing}, the system aims to present each user with climate change impacts and solutions that are most relevant and motivational based on that user's personal values. To achieve this, the system must 
a) access a knowledge base of climate change impacts and solutions b) be provided a user's motivational profile, and c) rank the climate concepts based on the user's profile. Figure~\ref{fig:pipeliine} shows the outline of the proposed recommendation system.

The schema for solutions in the knowledge base relies on having impacts standardized, therefore we focus our work (and this proposal) on curating impacts as {\em cause-effect} concepts before defining a climate solutions schema. Traditionally, domain experts would curate the knowledge base with these concepts from a set of climate science articles. However, manual curation is hard to scale and automatic curation techniques fail to generalize well to new domains~\cite{auer2007dbpedia,etzioni2008open,mintz2009distant}. In this work, we build a novel knowledge discovery system to build a climate knowledge base (ClimateKB) semi-automatically from the articles. Additionally, the climate concepts in ClimateKB are linked to motivational reference characteristics by domain experts since it is hard to mine these correlations in the absence of user interaction information. Lastly, we obtain a user motivational profile using a questionnaire and rank the concepts in ClimateKB based on the user's profile. To the best of our knowledge, this is the first initiative to build a ClimateKB and a climate recommendation system and revisits open research challenges in these problems. 

To summarize, our main contributions are: (1) a knowledge base, ClimateKB, that contains causes and effects of climate change, (2) a knowledge discovery system for populating ClimateKB semi-automatically from text, and (3) a recommendation system powered by ClimateKB. Since ClimateKB has many more potential downstream applications such as fact-checking and retrieval, we will freely release it in the easily accessible Web Ontology Language (OWL) format.

\section{Climate Knowledge Discovery System}\label{sec:kbc}

\begin{figure}[ht]
    \centering
    \includegraphics[scale=0.5]{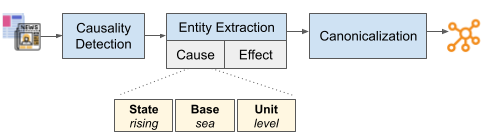}
    \caption{Knowledge discovery pipeline with 3 key components: causal sentence detection, entity extraction, and canonicalization.}
    \label{fig:pipeline}
\end{figure}

Figure~\ref{fig:pipeline} shows the overview of our knowledge discovery system. Given a corpus of trusted climate articles, the system first identifies sentences describing cause-effect relationships about climate change. It then finds entity mentions, the {\em cause} and the {\em effect}, from each causal sentence. It then canonicalizes the entity mentions and verifies the climate facts before populating the ClimateKB.

\subsection{ClimateKB}

\begin{figure}[ht]
    \centering
    \includegraphics[scale=0.52]{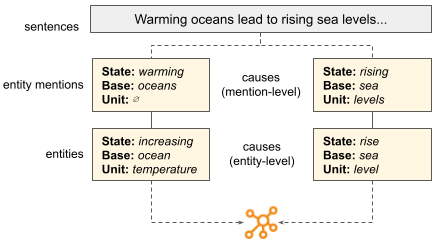}
    \caption{Data Model of ClimateKB}
    \label{fig:kb}
\end{figure}

We first describe the data model (Figure~\ref{fig:kb}) of ClimateKB. 
We focus on sentences in climate articles to find cause-effect relationships. An entity mention is a reference to a climate entity, such as the phrase \phrase{warming ocean} in the example sentence. A climate entity is a real-world concept relating to climate science (e.g., \phrase{sea level rise}), social science (e.g., \phrase{increased conflict events}), etc.  

In order to accurately represent information about climate change, we model each mention as a tuple of the form $(\textit{state}, \textit{base}, \textit{unit})$. For instance, simply extracting \phrase{ocean} or \phrase{sea} as the entity will lead to erroneous facts in the KB. We, therefore, additionally extract the associated state of change and the unit of measurement. Note that the vastness of entities in the domain makes ClimateKB unique from other general-purpose KBs about real-world entities (e.g., person, location) or concepts (e.g., drug names). 

\subsection{Data Collection}

We focus on news articles to build the ClimateKB and power the recommendation system. Unlike scientific articles that tend to contain complicated jargon, news articles are easy for users and domain experts to understand. Additionally, well-reputed news sources often cite scientific articles and summarize their key information more coherently. To build a corpus of reliable climate articles, we manually aggregated popular, reputed, and relevant news articles over a period of one year. Our final corpus has about 800 articles on a broad diversity of climate change issues including wildfires, coral bleaching, and extreme weather.

\subsection{Causality Detection}

The system next has to find causal sentences from the climate articles. While transformer-based models~\cite{bert} have shown state-of-the-art performance on the task~\cite{causalbert}, their performance is known to deteriorate substantially on out-of-domain datasets. To boost the performance, we first adapt the model to our domain by further pre-training BERT on climate news articles, scientific papers, IPCC reports, and books written for the public by climatologists. We refer to the domain-adapted model as ClimateBERT. Next, we fine-tune ClimateBERT for causality-detection using SemEval2007~\cite{girju2010knowledge} and SemEval2010~\cite{hendrickx2019semeval} benchmarks. On a test set of 600 sentences curated by domain experts from climate articles, the resulting model achieves 90\% precision and 28\% recall. Although the performance is passable for the downstream tasks, further improvements require techniques for robust domain adaptation and fine-tuning with limited data. 

\subsection{Entity Extraction and Canonicalization}

ClimateKB entities are complex and typically have a {\em base}, a {\em state}, and a {\em unit}. The base refers to the core climate or social concept (e.g, wind, ocean, suicide), the unit refers to the measurable aspect of the base (e.g, events, frequency, level), and the state describe the change in the unit (e.g., increasing, warming). Since these entities are domain specific and complex, off-the-shelf named entity recognition models cannot be directly used.

End-to-end neural models~\cite{zhanming2019} and pre-trained language models~\cite{bert} have state-of-the-art NER performance. However, fine-tuning these models requires high-quality annotated data. Our initial attempts to curate training data for entity extraction reveal several challenges, including ambiguity in labels,  the presence of implicit entities, and the use of anaphora. Some of these challenges are illustrated in Example~\ref{ex:entity_extraction}.  

\begin{exmp}\label{ex:entity_extraction}
\noindent
\normalfont{E1: In the sentence, {\em ``warmer temperatures lead to ...''}, the entity \phrase{air} is implied.}

\noindent
\normalfont{E2: In the sentence, {\em ``climate pressures can adversely impact resource availability ...''}, the token \phrase{pressure} can be a base or unit.}

\noindent
\normalfont{E2: In the sentence, {\em ``this can trigger a chain of ...''}, the token \phrase{this} refers to an entity from the previous sentence.}

\end{exmp}

Annotators who are not careful or lack background knowledge can potentially make label mistakes, which can negatively impact model training. This opens up new research avenues for developing novel frameworks that can handle mistakes in the training data and/or can help guide the annotators so they make fewer mistakes. 

\section{ClimateKB-based recommendations}\label{sec:rec}

Our goal is to use ClimateKB to catalog climate change impacts and solutions and recommend climate information that best motivates users to take action. Due to the novelty of the task, our proposed system profiles the user explicitly based on researched reference characteristics and leverages the manually curated associations of these characteristics to entities in the ClimateKB.

For reference characteristics, we use the personal values framework from Schwartz's theory of basic human values~\cite{schwartz12,sagiv17}. Following prior work~\cite{ding2016personalized,OPDENAKKER2015104,feduc2019}, we focus on the following 10 personal values $v_i$: {conformity}, { tradition}, { benevolence}, { universalism}, { self-direction}, { stimulation}, { stimulation}, { hedonism}, { achievement}, { power}, and { security}. 

 For obtaining a user's personal values, we used a slightly modified version of the Portrait Value Questionnaire (PVQ)\cite{schwartz2003pvq}. Specifically, we modified the ultra short 10 question version \cite{sandy2017pvq} to refer directly to users instead of requesting that users compare themselves to someone of the same gender. Each question assesses a different personal value and uses a 6-point Likert scale~\cite{joshi2015Likert} with values from "strongly disagree" to "strongly agree".

\begin{exmp}\label{ex:example_associations}
\noindent
\normalfont{\em{``decrease in population of moose available to hunt''}}

\normalfont{Positive association: {\em power}, {\em stimulation}, {\em hedonism} \\
Negative association: {\em universalism}\\
Neutral association with remaining values
}
\end{exmp}

Next, the entities in the ClimateKB must be linked to the 10 personal value characteristics. To ensure high quality associations, we ask domain experts to assign an association: positive, negative, or neutral, to each entity applicable in the KB. Example~\ref{ex:example_associations} shows a climate entity phrase and its associations to different characteristics of personal values. More formally, for a personal value ${v_i}$ they assign an association score $a_{v_i}$, where $a_{v_i}$ is $1$ if the association is positive, $-1$ if the association is negative and $0$ if the association is neutral. 

Lastly, we compute the relevance of a climate entity to a user. Note the scoring method proposed is for proof-of-concept and is simple.   
Let $u_{v_i}$ indicate the positive, centered, and scaled Likert score of a personal value $v_i$ obtained from a user's responses to the questionnaire. Given a climate entity $e$, let $a_{v_i}^e$ indicate the different associations of the entity. The relevance $S_e$ of climate entity $e$ to the user then can be computed by: 

\vspace*{-\abovedisplayskip}
\vspace*{-2em}
\begin{align*}
    S_e = \sum_{n=1}^{10} u_{v_i} \cdot a_{v_i}^e
\end{align*}
\vspace*{-\belowdisplayskip}
\vspace*{-1em}

The recommendation system could be improved to use more sophisticated measures that learn from user interactions in the app. More refined and sophisticated versions of our proposed motivation scoring and recommendation system are likely to exist, and could be evaluated by comparing each to actual user's behavior, preferences, and psychology.

\section{Future Outlook}
In the coming years, we expect climate change to worsen. Consequently, there will be more news coverage around those impacts and more need for climate action. This research will help us build a pipeline that expands the ClimateKB in a scalable manner. This research also builds the foundation for us to effectively capture and add to the ClimateKB climate change adaptations and solutions, even as more are developed. Expanding ClimateKB will bolster our recommendation system, allowing it to motivate a more diverse set of people and generate more conversation around climate change. Since our work is freely available, we hope others can find additional and novel applications of ClimateKB and other models we provide. 

\section*{Acknowledgments}
Cloud computing for this work is supported by an AI For Earth grant from Microsoft, with additional intellectual and volunteer support from Scientists Speak Up at Stanford University. We thank the current Climate Mind team beyond those listed as authors on this manuscript (in alphabetical order: Himesh Buch, Nick Callaghan, Alexis Carras, Kay Cochrane, Elle Dashfield, Yasmine Himanen, Veni Mittal, Stefanie Müller, Camille Naidoo, Henry Nguyen, Johan Olsson, Sean Payne, Rohan Wanchoo, and Lukas Ziegler). We are also very grateful for the help of Tycho Tax for advice on this proposal and many others who have or hope to contribute to the Climate Mind project. If interested in volunteering with our diverse, international team, please visit climatemind.org for information on how to join.

\bibliography{main}

\begin{thebibliography}{25}
\providecommand{\natexlab}[1]{#1}
\providecommand{\url}[1]{\texttt{#1}}
\expandafter\ifx\csname urlstyle\endcsname\relax
  \providecommand{\doi}[1]{doi: #1}\else
  \providecommand{\doi}{doi: \begingroup \urlstyle{rm}\Url}\fi

\bibitem[Akin \& Landrum(2017)Akin and Landrum]{akin2017recap}
Akin, H. and Landrum, A.
\newblock A recap: Heuristics, biases, values, and other challenges to
  communicating science.
\newblock \emph{The Oxford Handbook of the Science of Science Communication},
  455:\penalty0 460, 2017.

\bibitem[Auer et~al.(2007)Auer, Bizer, Kobilarov, Lehmann, Cyganiak, and
  Ives]{auer2007dbpedia}
Auer, S., Bizer, C., Kobilarov, G., Lehmann, J., Cyganiak, R., and Ives, Z.
\newblock Dbpedia: A nucleus for a web of open data.
\newblock In \emph{The semantic web}, pp.\  722--735. Springer, 2007.

\bibitem[Carpenter(2019)]{carpenter2019cognitive}
Carpenter, C.~J.
\newblock Cognitive dissonance, ego-involvement, and motivated reasoning.
\newblock \emph{Annals of the International Communication Association},
  43\penalty0 (1):\penalty0 1--23, 2019.

\bibitem[Chong \& Druckman(2007)Chong and Druckman]{chong2007framing}
Chong, D. and Druckman, J.~N.
\newblock Framing theory.
\newblock \emph{Annu. Rev. Polit. Sci.}, 10:\penalty0 103--126, 2007.

\bibitem[Devlin et~al.(2019)Devlin, Chang, Lee, and Toutanova]{bert}
Devlin, J., Chang, M.-W., Lee, K., and Toutanova, K.
\newblock {BERT}: Pre-training of deep bidirectional transformers for language
  understanding.
\newblock In \emph{Proceedings of the 2019 Conference of the North {A}merican
  Chapter of the Association for Computational Linguistics: Human Language
  Technologies, Volume 1 (Long and Short Papers)}, pp.\  4171--4186,
  Minneapolis, Minnesota, June 2019. Association for Computational Linguistics.
\newblock \doi{10.18653/v1/N19-1423}.
\newblock URL \url{https://www.aclweb.org/anthology/N19-1423}.

\bibitem[Ding \& Pan(2016)Ding and Pan]{ding2016personalized}
Ding, T. and Pan, S.
\newblock Personalized emphasis framing for persuasive message generation.
\newblock In \emph{Proceedings of the 2016 Conference on Empirical Methods in
  Natural Language Processing}, pp.\  1432--1441, Austin, Texas, November 2016.
  Association for Computational Linguistics.
\newblock \doi{10.18653/v1/D16-1150}.
\newblock URL \url{https://www.aclweb.org/anthology/D16-1150}.

\bibitem[Druckman \& McGrath(2019)Druckman and McGrath]{druckman2019evidence}
Druckman, J.~N. and McGrath, M.~C.
\newblock The evidence for motivated reasoning in climate change preference
  formation.
\newblock \emph{Nature Climate Change}, 9\penalty0 (2):\penalty0 111--119,
  2019.

\bibitem[Etzioni et~al.(2008)Etzioni, Banko, Soderland, and
  Weld]{etzioni2008open}
Etzioni, O., Banko, M., Soderland, S., and Weld, D.~S.
\newblock Open information extraction from the web.
\newblock \emph{Communications of the ACM}, 51\penalty0 (12):\penalty0 68--74,
  2008.

\bibitem[Girju et~al.(2010)Girju, Beamer, Rozovskaya, Fister, and
  Bhat]{girju2010knowledge}
Girju, R., Beamer, B., Rozovskaya, A., Fister, A., and Bhat, S.
\newblock A knowledge-rich approach to identifying semantic relations between
  nominals.
\newblock \emph{Information processing \& management}, 46\penalty0
  (5):\penalty0 589--610, 2010.

\bibitem[Hendrickx et~al.(2019)Hendrickx, Kim, Kozareva, Nakov, S{\'e}aghdha,
  Pad{\'o}, Pennacchiotti, Romano, and Szpakowicz]{hendrickx2019semeval}
Hendrickx, I., Kim, S.~N., Kozareva, Z., Nakov, P., S{\'e}aghdha, D.~O.,
  Pad{\'o}, S., Pennacchiotti, M., Romano, L., and Szpakowicz, S.
\newblock Semeval-2010 task 8: Multi-way classification of semantic relations
  between pairs of nominals.
\newblock \emph{arXiv preprint arXiv:1911.10422}, 2019.

\bibitem[Howe et~al.(2015)Howe, Mildenberger, Marlon, and
  Leiserowitz]{howe2015geographic}
Howe, P.~D., Mildenberger, M., Marlon, J.~R., and Leiserowitz, A.
\newblock Geographic variation in opinions on climate change at state and local
  scales in the usa.
\newblock \emph{Nature climate change}, 5\penalty0 (6):\penalty0 596--603,
  2015.

\bibitem[Iyengar \& Massey(2019)Iyengar and Massey]{iyengar2019scientific}
Iyengar, S. and Massey, D.~S.
\newblock Scientific communication in a post-truth society.
\newblock \emph{Proceedings of the National Academy of Sciences}, 116\penalty0
  (16):\penalty0 7656--7661, 2019.

\bibitem[Jie \& Lu(2019)Jie and Lu]{zhanming2019}
Jie, Z. and Lu, W.
\newblock Dependency-guided lstm-crf for named entity recognition.
\newblock In \emph{Proceedings of the 2019 Conference on Empirical Methods in
  Natural Language Processing and the 9th International Joint Conference on
  Natural Language Processing}, pp.\  3862–--3872, 2019.

\bibitem[Joshi et~al.(2015)Joshi, Kale, Chandel, and Pal]{joshi2015Likert}
Joshi, A., Kale, S., Chandel, S., and Pal, D.~K.
\newblock Likert scale: Explored and explained.
\newblock \emph{Current Journal of Applied Science and Technology}, pp.\
  396--403, 2015.

\bibitem[Khetan et~al.(2021)Khetan, Ramnani, Anand, Sengupta, and
  Fano]{causalbert}
Khetan, V., Ramnani, R., Anand, M., Sengupta, S., and Fano, A.~E.
\newblock Causal bert : Language models for causality detection between events
  expressed in text, 2021.

\bibitem[Leiserowitz~A. \& E.(2021)Leiserowitz~A. and E.]{6americasReview}
Leiserowitz~A., Roser-Renouf~C., M.~J. and E., M.
\newblock Global warming’s six americas: a review and recommendations for
  climate change communication.
\newblock \emph{urrent Opinion in Behavioral Sciences}, 42:\penalty0 97--103,
  2021.
\newblock \doi{10.1016/j.cobeha.2021.04.007}.

\bibitem[Leuzinger et~al.(2019)Leuzinger, Borrelle, and Jarvis]{feduc2019}
Leuzinger, S., Borrelle, S.~B., and Jarvis, R.~M.
\newblock Improving climate-change literacy and science communication through
  smart device apps.
\newblock \emph{Frontiers in Education}, 4:\penalty0 138, 2019.
\newblock ISSN 2504-284X.
\newblock \doi{10.3389/feduc.2019.00138}.
\newblock URL
  \url{https://www.frontiersin.org/article/10.3389/feduc.2019.00138}.

\bibitem[Masson-Delmotte et~al.(2018)Masson-Delmotte, Zhai, Pörtner, Roberts,
  Skea, Shukla, Pirani, Moufouma-Okia, Péan, Pidcock, Connors, Matthews, Chen,
  Zhou, Gomis, Lonnoy, Maycock, Tignor, and Waterfield]{ipccsr15}
Masson-Delmotte, V., Zhai, P., Pörtner, H.-O., Roberts, D., Skea, J., Shukla,
  P., Pirani, A., Moufouma-Okia, W., Péan, C., Pidcock, R., Connors, S.,
  Matthews, J., Chen, Y., Zhou, X., Gomis, M., Lonnoy, E., Maycock, T., Tignor,
  M., and Waterfield, T.
\newblock Global warming of 1.5°c. an ipcc special report on the impacts of
  global warming of 1.5°c above pre-industrial levels and related global
  greenhouse gas emission pathways, in the context of strengthening the global
  response to the threat of climate change, sustainable development, and
  efforts to eradicate poverty.
\newblock Technical report, International Panel on Climate Change, 2018.

\bibitem[Mintz et~al.(2009)Mintz, Bills, Snow, and Jurafsky]{mintz2009distant}
Mintz, M., Bills, S., Snow, R., and Jurafsky, D.
\newblock Distant supervision for relation extraction without labeled data.
\newblock In \emph{Proceedings of the Joint Conference of the 47th Annual
  Meeting of the ACL and the 4th International Joint Conference on Natural
  Language Processing of the AFNLP}, pp.\  1003--1011, 2009.

\bibitem[{op den Akker} et~al.(2015){op den Akker}, Cabrita, {op den Akker},
  Jones, and Hermens]{OPDENAKKER2015104}
{op den Akker}, H., Cabrita, M., {op den Akker}, R., Jones, V.~M., and Hermens,
  H.~J.
\newblock Tailored motivational message generation: A model and practical
  framework for real-time physical activity coaching.
\newblock \emph{Journal of Biomedical Informatics}, 55:\penalty0 104--115,
  2015.
\newblock ISSN 1532-0464.
\newblock \doi{https://doi.org/10.1016/j.jbi.2015.03.005}.
\newblock URL
  \url{https://www.sciencedirect.com/science/article/pii/S1532046415000489}.

\bibitem[Sagiv et~al.(2017)Sagiv, Roccas, Cieciuch, and Schwartz]{sagiv17}
Sagiv, L., Roccas, S., Cieciuch, J., and Schwartz, S.~H.
\newblock Personal values in human life.
\newblock In \emph{Nature human behaviour, 1(9)}, pp.\  630--639, 2017.
\newblock \doi{10.1038/s41562-017-0185-3}.
\newblock URL \url{https://doi.org/10.1038/s41562-017-0185-3}.

\bibitem[Sandy et~al.(2017)Sandy, Gosling, Schwartz, and
  Koelkebeck]{sandy2017pvq}
Sandy, C.~J., Gosling, S.~D., Schwartz, S.~H., and Koelkebeck, T.
\newblock The development and validation of brief and ultrabrief measures of
  values.
\newblock \emph{Journal of Personality Assessment}, 99\penalty0 (5):\penalty0
  545--555, 2017.
\newblock \doi{10.1080/00223891.2016.1231115}.
\newblock URL \url{https://doi.org/10.1080/00223891.2016.1231115}.
\newblock PMID: 27767342.

\bibitem[Scheufele(2014)]{scheufele14}
Scheufele, D.~A.
\newblock Science communication as political communication.
\newblock \emph{Proceedings of the National Academy of Sciences}, 111\penalty0
  (Supplement 4):\penalty0 13585--13592, 2014.
\newblock \doi{10.1073/pnas.1317516111}.
\newblock URL \url{https://www.pnas.org/content/111/Supplement_4/13585}.

\bibitem[Schwartz(2003)]{schwartz2003pvq}
Schwartz, S.
\newblock A proposal for measuring value orientations across nations.
\newblock \emph{Questionnaire Package of ESS}, pp.\  259--290, 01 2003.

\bibitem[Schwartz(2012)]{schwartz12}
Schwartz, S.~H.
\newblock An overview of the schwartz theory of basic values.
\newblock In \emph{Online Readings in Psychology and Culture, 2(1)}, 2012.
\newblock \doi{10.9707/2307-0919.1116}.
\newblock URL \url{https://doi.org/10.9707/2307-0919.1116}.

\end{thebibliography}
\bibliographystyle{icml2021}

\end{document}